\documentclass[11pt]{article}
\usepackage{flushend}
\usepackage[]{acl2022}
\usepackage{times}
\usepackage{latexsym}

\usepackage{microtype}
\usepackage{graphicx}
\usepackage{booktabs}  
\usepackage{multirow}
\usepackage{subfigure}
\usepackage{verbatim}
\usepackage{CJKutf8} 
\usepackage{makecell}
\usepackage{amsmath}
\usepackage{amssymb}
\usepackage{listings}
\DeclareFixedFont{\ttb}{T1}{txtt}{bx}{n}{8} % for bold
\DeclareFixedFont{\ttm}{T1}{txtt}{m}{n}{8}  % for normal

\usepackage{color}
\definecolor{deepblue}{rgb}{0,0,0.5}
\definecolor{deepred}{rgb}{0.6,0,0}
\definecolor{deepgreen}{rgb}{0,0.5,0}

\newcommand\pythonstyle{\lstset{
language=Python,
basicstyle=\ttm,
morekeywords={self},              % Add keywords here
keywordstyle=\ttb\color{deepblue},
emph={MyClass,__init__},          % Custom highlighting
emphstyle=\ttb\color{deepred},    % Custom highlighting style
stringstyle=\color{deepgreen},
frame=tb,                         % Any extra options here
showstringspaces=false
}}

\lstnewenvironment{python}[1][]
{
\pythonstyle
\lstset{#1}
}
{}

\newcommand\pythoninline[1]{{\pythonstyle\lstinline!#1!}}

\title{NoisyTune: A Little Noise Can Help You Finetune \\Pretrained Language Models Better}

\author{Chuhan Wu$^\dagger$~~~~Fangzhao Wu$^\ddagger$\thanks{~~Corresponding author.}~~~~Tao Qi$^\dagger$~~~~Yongfeng Huang$^\dagger$~~~~Xing Xie$^\ddagger$\\
    $^\dagger$Department of Electronic Engineering, Tsinghua University, Beijing 100084, China  \\
     $^\ddagger$Microsoft Research Asia, Beijing 100080, China\\
  {\tt\{wuchuhan15, wufangzhao, taoqi.qt\}@gmail.com} \\ {\tt yfhuang@tsinghua.edu.cn, xingx@microsoft.com}
  }
\date{}

\begin{document}
\maketitle

\begin{abstract}

Effectively finetuning pretrained language models (PLMs) is critical for their success in downstream tasks.
However, PLMs may have risks in overfitting the pretraining tasks and data, which usually have gap with the target downstream tasks.
Such gap may be difficult for existing PLM finetuning methods to overcome and lead to suboptimal performance.
In this paper, we propose a very simple yet effective method named \textit{NoisyTune} to help better finetune PLMs on downstream tasks by adding some noise to the parameters of PLMs before finetuning.
More specifically, we propose a matrix-wise perturbing method which adds different uniform noises to different parameter matrices based on their standard deviations. In this way, the varied characteristics of different types of parameters in PLMs can be considered.
Extensive experiments on both GLUE English benchmark and XTREME multilingual benchmark show \textit{NoisyTune} can consistently empower the finetuning of different PLMs on different downstream tasks.

\end{abstract}

\section{Introduction}

In recent years, pretrained language models (PLMs)  have achieved huge success in NLP~\cite{qiu2020pre}.
Many PLMs such as BERT~\cite{devlin2019bert}, RoBERTa~\cite{liu2019roberta} and UniLM~\cite{dong2019unified} which are pretrained from large-scale unlabeled corpus in a self-supervised way, have significantly improve various downstream tasks such as reading comprehension~\cite{xu2019bert}, machine translation~\cite{brown2020language}, text classification~\cite{bao2020unilmv2}, dialog~\cite{wu2020tod} and recommendation~\cite{wu2021plm} by finetuning on these tasks.

How to effectively finetune PLMs to better empower downstream tasks is an important research problem~\cite{zheng2021con}.
Many existing NLP methods usually directly finetune PLMs with the labeled data in downstream tasks~\cite{sun2019fine}.
Only a few works explore more effective and robust PLM finetuning methods~\cite{chen2020recall,Lee2020mixout,aghajanyan2020better,zhang2021revisiting,xu2021raise}.
For example, \citet{chen2020recall} proposed RecAdam that adds a penalty item to minimize the $L_2$ distance between the fine-tuned models and the pretrained models, where the penalty intensity is time-variant during finetuning.
\citet{Lee2020mixout} proposed Mixout which randomly replaces part of the parameters in the finetuned model with their original weights in the PLMs.
These PLM finetuning methods mainly focus on preventing PLMs from overfitting the limited labeled data in downstream tasks.
Besides the overfitting of downstream task data, a rarely studied problem is that the PLMs usually overfit the pretraining tasks and data~\cite{qi2020prophetnet}, which may have significant gap with the downstream task and data.
It is not easy for existing PLM finetuning methods to overcome such gap~\cite{roberts2020much}, which may lead to suboptimal performance especially when labeled data in downstream tasks is insufficient.

\begin{figure}[t]
  \centering
    \includegraphics[width=0.48\textwidth]{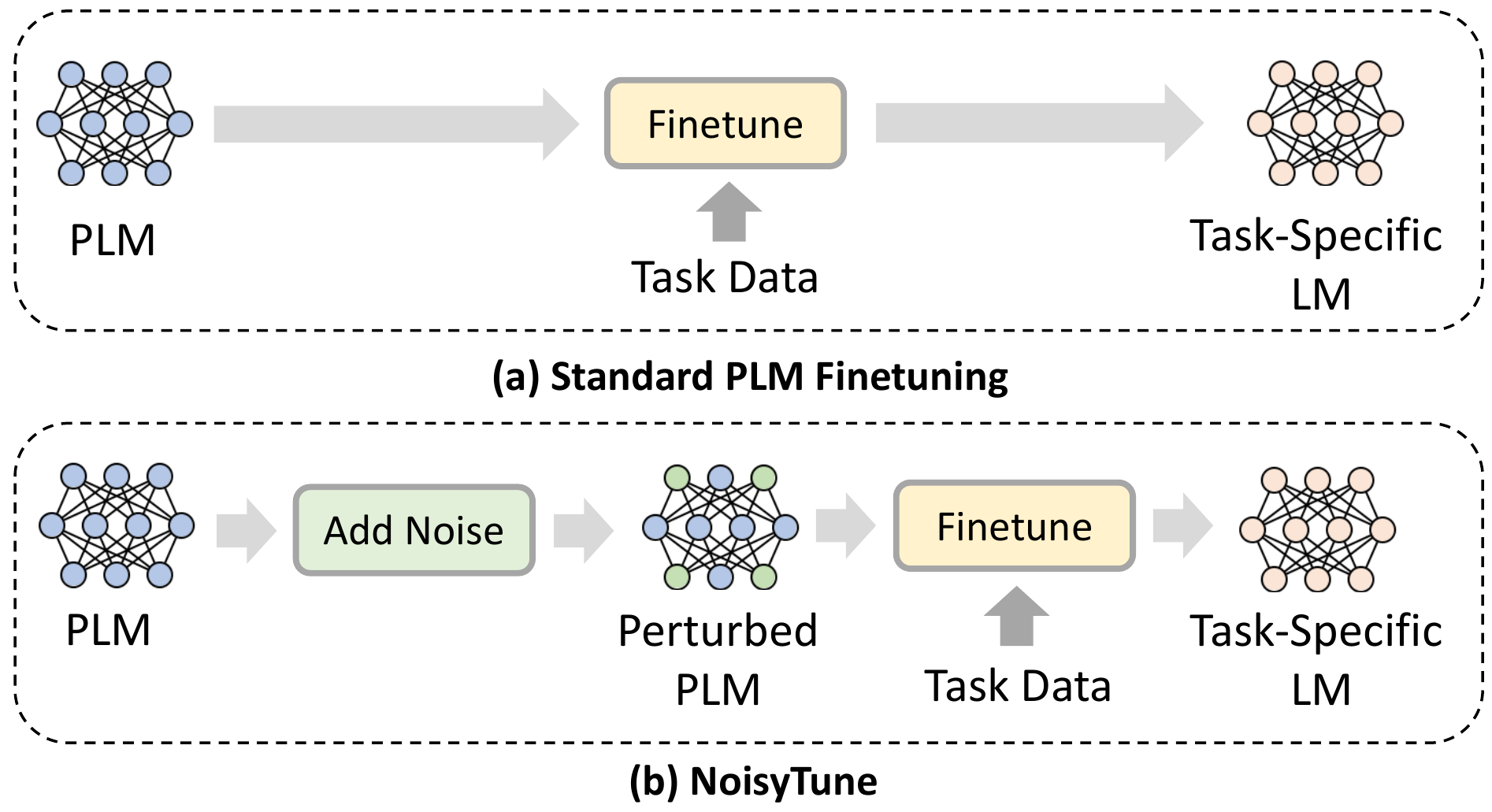}

  \caption{Schematic comparisons between standard PLM finetuning and our \textit{NoisyTune}.}\label{fig.exp}

\end{figure}

%Noisy training is a widely used technique to mitigate overfitting and learn more robust models~\cite{lecun2015deep}.
%A most common way is adding noise to the input to generate augmented training data~\cite{bishop1995training}, which has been proven effective in achieving model regularization abilities~\cite{hua2021noise}. 
%Researchers have also studied adding noise to weights~\cite{jim1996analysis}, gradients~\cite{smilkov2017smoothgrad}, and outputs targets~\cite{goodfellow2014explaining}.
%However, these works mainly focus on using noise as a ``model regularizer'' to avoid overfitting in downstream tasks, while they may not be optimal in tackling the problem of overfitting self-supervision tasks.

In order to handle this problem, in this paper we propose a very simple yet effective method named \textit{NoisyTune}, which can help better finetune PLMs for downstream tasks.
Different from the standard finetuning paradigm (Fig.~\ref{fig.exp} (a)) which directly finetunes PLMs on the downstream task data, the key idea of \textit{NoisyTune} is to add a small amount of noise to perturb PLMs parameters before finetuning (Fig.~\ref{fig.exp} (b)).
It can help prevent PLMs from overfitting the tasks and data in the pretraining stage, and reduce the gap between pretraining and downstream tasks.
Since PLMs have different types of parameters which usually own different characteristics, in \textit{NoisyTune} we use a matrix-wise perturbing method that adds uniform noise with different intensities to different parameter matrices according to their standard deviations for better adaptation.
We conduct extensive experiments on two widely used NLP benchmarks, namely, GLUE~\cite{wang2018glue} for English language understanding and XTREME~\cite{hu2020xtreme} for multilingual language understanding.
The results show \textit{NoisyTune} can empower the finetuning of different PLMs on many different downstream NLP tasks to consistently achieve better performance.
In addition, the results show \textit{NoisyTune} can be easily combined with many existing PLM finetuning methods and further improve their performance.

\section{NoisyTune}

The goal of \textit{NoisyTune} is for more effective  finetuning of PLMs on downstream tasks.
The motivation of \textit{NoisyTune} is that PLMs are well pretrained on some unlabeled corpus with some self-supervision tasks, and they may overfit these pretraining data and tasks~\cite{qi2020prophetnet}, which usually have gap with the downstream task and data.
It may be difficult for PLMs to effectively adapt to downstream tasks especially when labeled data in these tasks are limited, which is usually the case.
Motivated by the dueling bandits mechanism~\cite{yue2009interactively} that adds randomness to the model for exploration, as shown in Fig.~\ref{fig.exp}, we propose to add some noise to the parameters of PLMs before finetuning them on downstream tasks to do some ``exploration'' in parameter space and reduce the risk of overfitting the pretraining tasks and data.

PLMs usually have different kinds of parameter matrices, such as query, key, value, and feedforward network matrices~\cite{devlin2019bert}.
Different parameter matrices in the PLMs usually have different characteristics and scales.
For example, some researchers found that the self-attention parameters and the feed-forward network parameters in Transformers have very different properties, such as rank and density~\cite{wang2020linformer}.
Thus, adding unified noise to all parameter matrices in PLMs may not be optimal for keeping their good model utility.
To handle this challenge, we propose a matrix-wise perturbing method that adds  noise with different intensities to different parameter matrices according to their variances.
Denote the parameter matrices (or scalars/vectors) in a PLM as $[\mathbf{W}_1, \mathbf{W}_2, ..., \mathbf{W}_N]$, where $N$ is the number of parameter matrix types.
Denote the perturbed version of the parameter matrix $\mathbf{W}_i$ as $\tilde{\mathbf{W}}_i$, which is computed as follows:
\begin{equation}
    \tilde{\mathbf{W}}_i=\mathbf{W}_i+ U(-\frac{\lambda}{2}, \frac{\lambda}{2})*std(\mathbf{W}_i),
\end{equation}
where $\rm{std}$ stands for standard deviation. The function $U(a,b)$ represents uniform distribution noise ranged from $a$ to $b$, and $\lambda$ is a hyperparameter that controls the relative noise intensity.\footnote{Note that $U(a,b)$ is a matrix with the same shape with $\mathbf{W}_i$ rather than a scalar.}
We can see that in \textit{NoisyTune} parameters in PLMs with higher variance will be added with stronger noise.
In addition, in some PLMs there are some constant matrices, such as token type embeddings in RoBERTa~\cite{liu2019roberta}.
They will not be perturbed because their standard deviation is 0.
It can ensure that these constant matrices will not be accidentally activated by additional noise.

\textit{NoisyTune} is a simple and general plug-and-play technique that can be applied to the finetuning of any PLM on any task, simply by inserting the following PyTorch-style code before finetuning:
\begin{python} 
for name,para in model.named_parameters():
    model.state_dict[name][:] +=
    (torch.rand(para.size())-0.5)
    *noise_lambda*torch.std(para) 
\end{python}

\section{Experiments}

\subsection{Datasets and Experimental Settings}

We conduct extensive experiments on two widely used benchmarks for PLM evaluation.
The first one is GLUE~\cite{wang2018glue}, which is a  benchmark for English language understanding that contains different tasks like natural language inference, sentiment analysis and sentence similarity evaluation.
The second one is XTREME~\cite{hu2020xtreme}, which is a benchmark for multilingual language understanding.
It covers  40 languages and contains four groups of tasks, including sentence classification, structured prediction, sentence retrieval and question answering.
More details of these benchmarks can refer to their original papers and official websites.
Since the test labels of GLUE are not released, following~\cite{bao2020unilmv2} we report results on the dev set of GLUE. 
The XTREME results are evaluated on the test set.
The hyperparameter $\lambda$ is 0.15 on GLUE and is 0.1 on XTREME.
The searching range of hyperparameters in our work are listed in Table~\ref{hyper}.

\begin{table}[h]
\resizebox{1.0\linewidth}{!}{
\begin{tabular}{lc}
\hline
\textbf{Hyperparameters} & \textbf{Range}                         \\ \hline
Learning rate            & \{7e-6, 1e-5, 2e-5, 3e-5\}             \\
Epoch                    & \{3, 5, 7, 10, 15, 20\}                \\
Batch size               & \{8, 16, 32\}                          \\
Noisy intensity          & \{0, 0.05, 0.1, 0.15, 0.2, 0.25, 0.3\} \\ \hline
\end{tabular}
}
\caption{Searching ranges of different hyperparameters in our experiments.} \label{hyper}
\end{table}

\begin{table*}[t]

\resizebox{1.0\linewidth}{!}{
\begin{tabular}{lcccccccccc}
\Xhline{1pt}
\multicolumn{1}{l}{\multirow{2}{*}{\textbf{Model}}} & \textbf{MNLI} & \textbf{QNLI} & \textbf{QQP} & \textbf{RTE} & \textbf{SST} & \textbf{MRPC} & \textbf{CoLA} & \textbf{STS} & \textbf{WNLI} & \textbf{} \\
\multicolumn{1}{c}{}                                & Acc           & Acc           & Acc          & Acc          & Acc          & Acc           & MCC           & PCC          & Acc           & Avg.      \\ \hline
BERT                                                & 84.4          & 91.5          & 90.9         & 67.7         & 93.0         & 87.1          & 58.1          & 89.4         & 54.4          & 79.6      \\
BERT+NoisyTune                                      & 84.7          & 91.8          & 91.2         & 68.8         & 93.4         & 88.0          & 59.0          & 90.1         & 56.1          & 80.3      \\ \hline
XLNET                                               & 86.6          & 91.6          & 91.2         & 72.9         & 94.4         & 88.1          & 59.6          & 89.6         & 57.5          & 81.3      \\
XLNET+NoisyTune                                     & 86.9          & 91.9          & 91.4         & 73.8         & 94.7         & 88.6          & 60.1          & 90.0         & 58.6          & 81.8      \\ \hline
RoBERTa                                             & 87.5          & 92.7          & 91.7         & 77.1         & 94.5         & 90.1          & 62.9          & 90.8         & 59.2          & 82.9      \\
RoBERTa+NoisyTune                                   & 87.8          & 93.1          & 91.9         & 78.8         & 94.9         & 90.6          & 63.6          & 91.1         & 60.3          & 83.6      \\ \hline
ELECTRA                                             & 88.4          & 92.9          & 91.7         & 75.2         & 94.9         & 88.2          & 64.2          & 90.1         &  62.0          & 83.1      \\
ELECTRA+NoisyTune                                   & 88.7          & 93.2          & 92.1         & 76.4         & 95.2         & 88.7          & 64.9          & 90.5         & 63.4          & 83.7      \\

\Xhline{1pt}
\end{tabular}
}
\caption{Results of different methods on the GLUE dev set.} \label{table.performance} 
\end{table*}

\begin{table*}[t]

\resizebox{1.0\linewidth}{!}{
\begin{tabular}{lcccccccccc}
\Xhline{1pt}
\multirow{2}{*}{\textbf{Model}} & \multicolumn{2}{c}{\textbf{Sentence Pair}} & \multicolumn{2}{c}{\textbf{Structured Prediction}} & \multicolumn{2}{c}{\textbf{Sentence Retrieval}} & \multicolumn{3}{c}{\textbf{Question Answering}} & \textbf{} \\
                                & XNLI                & PAWS-X               & POS                      & NER                     & BUCC                  & Tatoeba                 & XQuAD          & MLQA           & TyDiQA        &           \\ \hline
\textbf{Metrics}                & Acc                 & Acc                  & F1                       & F1                      & Acc                   & Acc                     & F1/EM          & F1/EM          & F1/EM         & Avg.      \\ \hline
\multicolumn{11}{l}{\textit{Fine-tune multilingual model on English training set (Cross-lingual Transfer)}}                                                                                                                                                \\ \hline
XLM-R$_{\rm{base}}$                       & 74.8                & 84.8                 & 75.5                     & 61.6                    & 77.6                  & 73.8                    & 71.9/56.6      & 65.2/47.0      & 55.5/38.4     & 70.0      \\
XLM-R$_{\rm{base}}$+NoisyTune             & 75.2                & 85.1                 & 76.0                     & 62.1                    & 78.2                  & 74.5                    & 72.3/57.1      & 65.5/47.4      & 56.0/39.2     & 70.5      \\ \hline
XLM-R$_{\rm{large}}$                      & 79.0                & 86.3                 & 72.7                     & 62.3                    & 79.2                  & 76.0                    & 76.2/60.4      & 71.4/53.0      & 65.0/45.0     & 72.4      \\
XLM-R$_{\rm{large}}$+NoisyTune            & 79.3                & 86.5                 & 73.5                     & 63.2                    & 79.9                  & 76.8                    & 76.7/61.0      & 71.9/53.6      & 65.4/45.6     & 73.0      \\ \hline
\multicolumn{11}{l}{\textit{Fine-tune multilingual model on all training sets (Translate-Train-All)}}                                                                                                                                                      \\ \hline
XLM-R$_{\rm{base}}$                       & 78.5                & 88.2                 & 76.2                     & 62.6                    & 79.6                  & 79.4                    & 75.0/61.5      & 67.8/50.1      & 63.8/47.6     & 73.3      \\
XLM-R$_{\rm{base}}$+NoisyTune             & 78.9                & 88.6                 & 76.8                     & 63.1                    & 80.0                  & 79.8                    & 75.4/61.8      & 68.0/50.4      & 64.1/48.1     & 73.7      \\ \hline
XLM-R$_{\rm{large}}$                      & 82.3                & 90.3                 & 77.3                     & 67.3                    & 82.5                  & 82.7                    & 80.0/65.6      & 72.9/54.4      & 66.3/47.6     & 76.4      \\
XLM-R$_{\rm{large}}$+NoisyTune            & 82.5                & 90.5                 & 77.8                     & 67.9                    & 82.9                  & 83.0                    & 80.4/66.1      & 73.3/54.9      & 66.8/48.2     & 76.8      \\ \Xhline{1pt}
\end{tabular}
}

\caption{Results of different methods on the XTREMRE test set.}\label{table.performance2} 
\end{table*}

Following~\cite{zheng2021con}, in sentence retrieval tasks we first train the models on the XNLI dataset, and then use the average of token representations produced by the hidden layer that yields the best performance.
In order not to harm the alignment of token embeddings across different languages, we do not add noise to the token embeddings in multilingual PLMs.
We repeat experiments 5 times with different random seeds and report the average scores.

\subsection{Performance Evaluation}

On the GLUE benchmark, we compare the performance of directly finetuning the base version of BERT~\cite{devlin2019bert}, XLNET~\cite{yang2019xlnet}, RoBERTa~\cite{liu2019roberta} and ELECTRA~\cite{clark2020electra} with that of finetuning them after applying \textit{NoisyTune}.
On the XTREME benchmark, we compare the performance of directly finetuning both base and large versions of XLM-R~\cite{conneau2020unsupervised} with that of their variants obtained by applying \textit{NoisyTune}.
The results on these two benchmarks are shown in Tables~\ref{table.performance} and~\ref{table.performance2}, respectively.
On the XTREME datasets, we report two types of results.
The first one is zero-shot crosslingual transfer from English to other languages, and the second one is learning models on both English and translated data.

According to these results, \textit{NoisyTune} can consistently improve the performance of different PLMs on different tasks in both English and multilingual settings.
In addition, the performance improvement brought by \textit{NoisyTune} is usually larger on relatively small datasets (e.g., RTE, CoLA and WNLI).
These results indicate that when labeled data in downstream tasks is insufficient, it is quite difficult to effectively finetune PLMs starting from the original parameters which usually overfit the pretraining tasks and data.
The experimental results validate that \textit{NoisyTune} can properly perturb PLMs with a little noise to explore different parameter spaces and reduce the overfitting problem, making PLMs easier to be adapted to downstream tasks.

\subsection{Which Noise to Use and How?}

\begin{figure}[!t]
  \centering
    \includegraphics[width=0.46\textwidth]{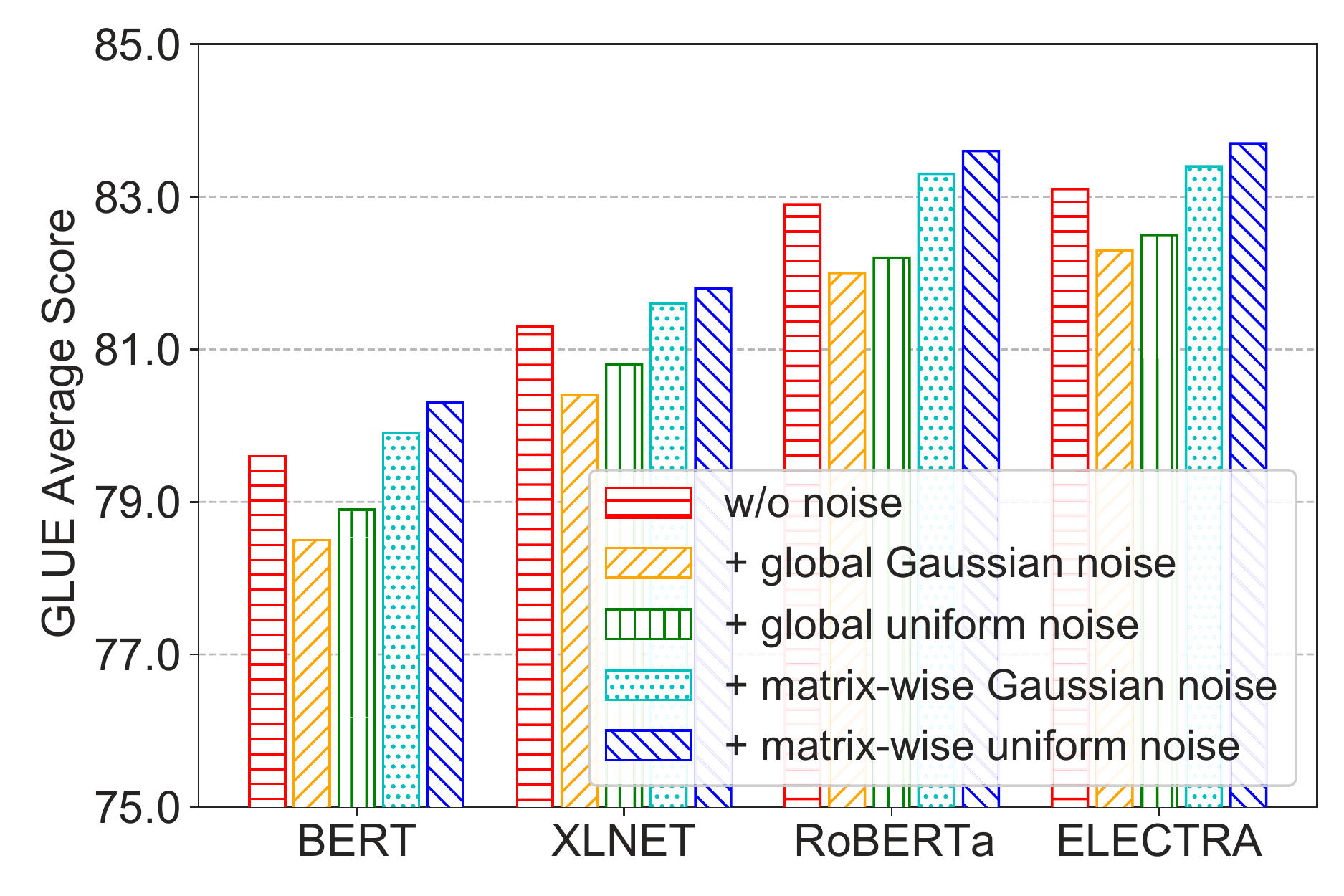}

  \caption{Different noise types and perturbing methods.}\label{fig.type}

\end{figure}

In this section we study which kind of noise is more suitable for \textit{NoisyTune}.
In addition, we explore whether our proposed matrix-wise perturbing method is better than using a unified global noise for all model parameters in PLMs. 
We compare five methods, including (1) \textit{NoisyTune} without any noise; (2) \textit{NoisyTune} with a global Gaussian noise; (3) \textit{NoisyTune} with a global uniform noise; (4) \textit{NoisyTune} with matrix-wise Gaussian noise; (5) \textit{NoisyTune} with matrix-wise uniform noise.
The results on GLUE are shown in Fig.~\ref{fig.type}, and the results on XTREME show similar patterns.
We find that adding global noise with the same distribution to all the PLM parameters will harm the model performance.
This is because different parameter matrices in PLMs have very different  distributions and characteristics~\cite{wang2020linformer}.
Simply adding a unified global noise to all the parameter matrices is not optimal.
The results show that matrix-wise noise is a much better choice, since the different characteristics of different parameter matrices can be taken into consideration.
In addition, we find an interesting phenomenon that adding uniform noise is better than Gaussian noise.
This may be because Gaussian noise has wider ranges and some extreme values may affect the model performance.
Thus, we use matrix-wise uniform noise in \textit{NoisyTune}.

\subsection{Combination with Existing PLM Finetuning Methods}

\begin{figure}[!t]
  \centering
    \includegraphics[width=0.46\textwidth]{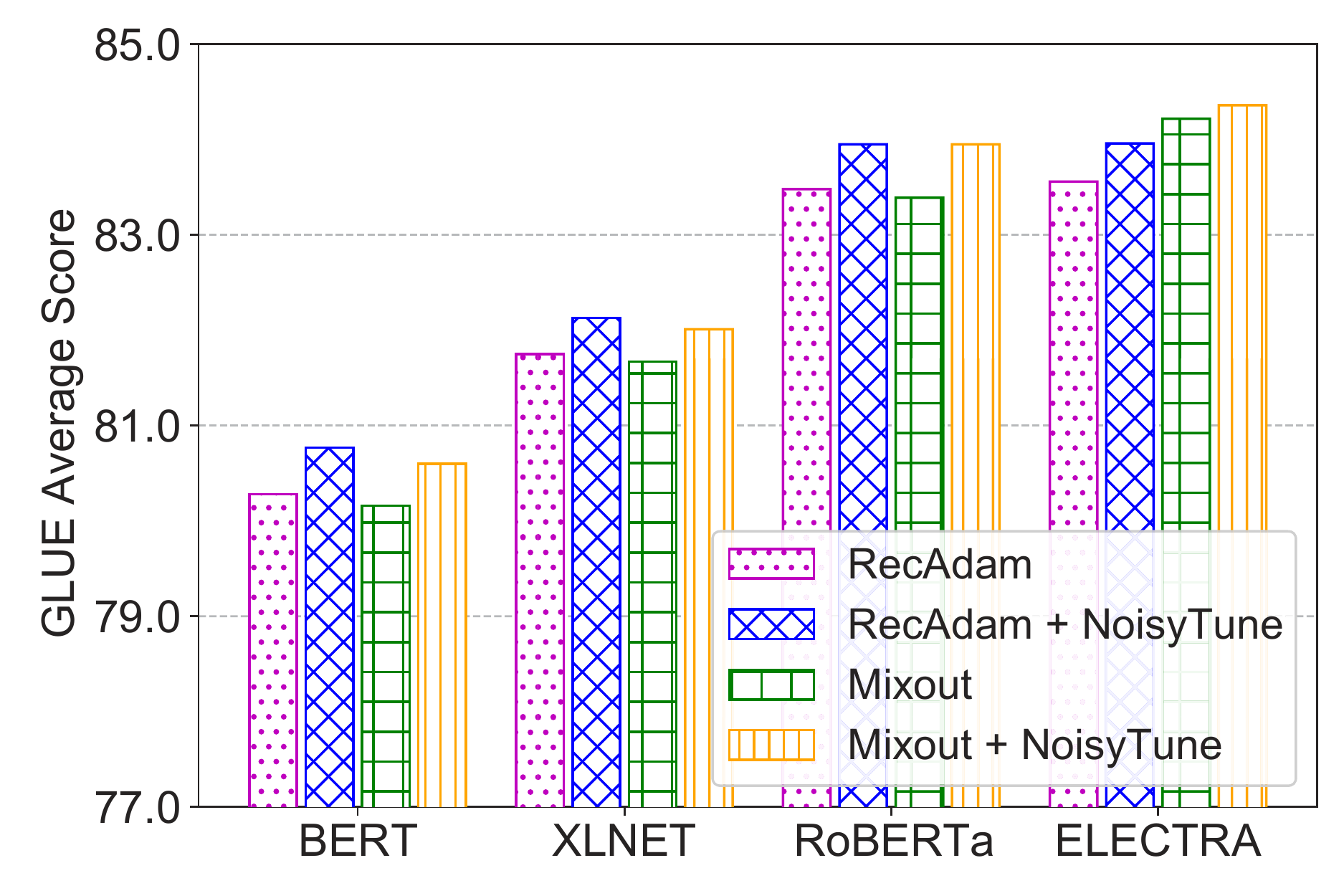}

  \caption{\textit{NoisyTune} can empower many existing PLM finetuning methods to achieve better performance.}\label{fig.combine}

\end{figure}

From Fig.~\ref{fig.exp}, it is very clear that \textit{NoisyTune} is independent of the specific PLM finetuning method, since it is applied at the stage before finetuning PLM on the task-specific data.
Thus, it is very easy to combine \textit{NoisyTune} with any kind of existing PLM finetuning method.
In this section, we explore whether \textit{NoisyTune} has the potential to empower the existing PLM finetuning techniques to achieve better performance.
Here we select two well-known PLM finetuning for experiments, i.e., RecAdam~\cite{chen2020recall} and Mixout~\cite{Lee2020mixout}.
The experimental results are summarized in Fig.~\ref{fig.combine}.
We find that combining \textit{NoisyTune} with existing PLM finetuning techniques can further improve their performance.
This is because \textit{NoisyTune} aims to address the overfitting of pretraining signals while these methods aim to prevent overfitting in downstream tasks. 
Thus, \textit{NoisyTune}  and these PLM finetuning methods are complementary, and they can be empowered by \textit{NoisyTune} to achieve better performance.

\subsection{Empirical Analysis of NoisyTune}

\begin{figure}[!t]
  \centering
    \includegraphics[width=0.46\textwidth]{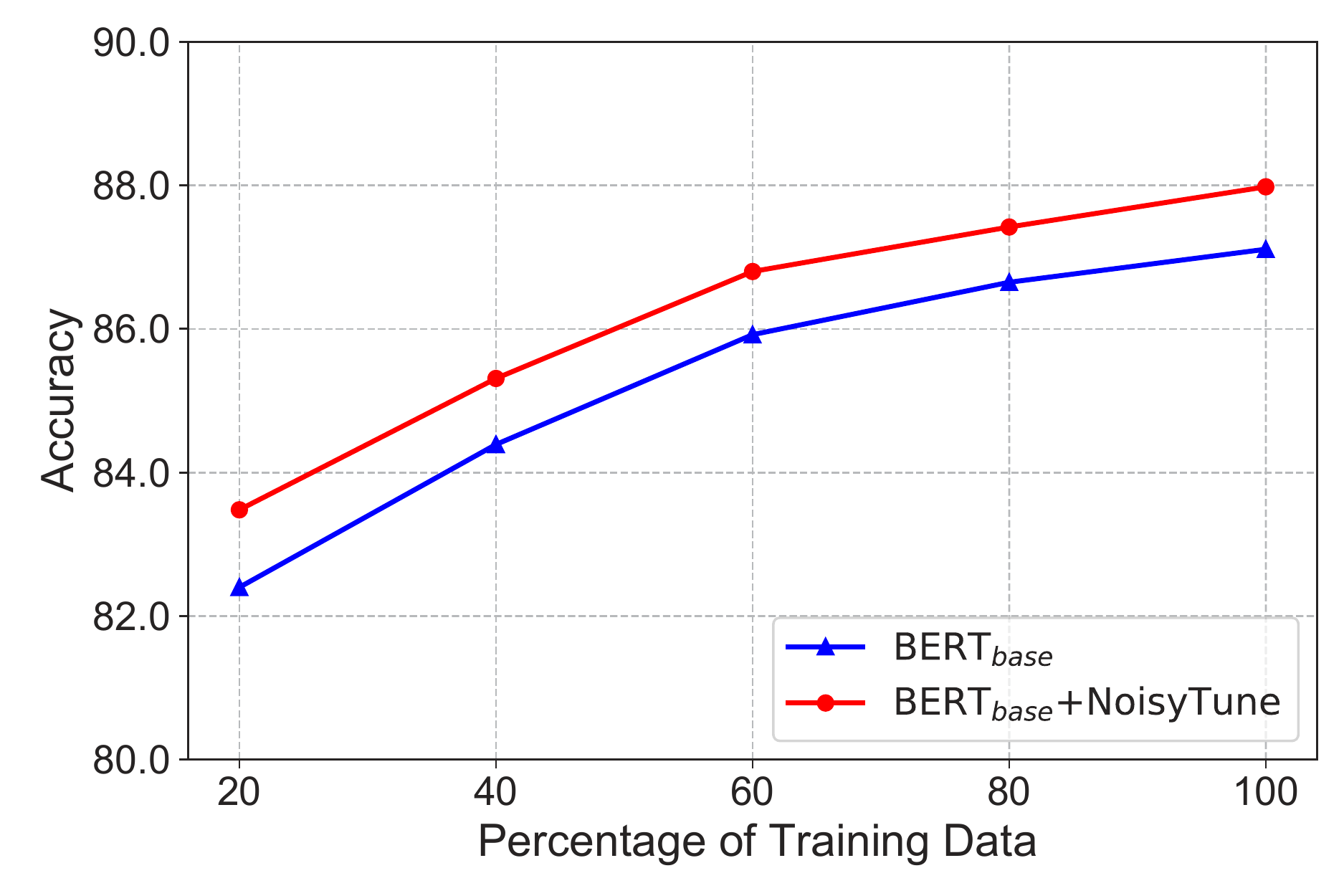}

  \caption{Influence of \textit{NoisyTune} on finetuning.}\label{fig.conv}

\end{figure}

Next, we empirically analyze why \textit{NoisyTune} can help PLM finetuning.
We compare the accuracy of BERT with and without \textit{NoisyTune} finetuned with different percentage of samples on the MRPC dataset.\footnote{We observe similar patterns on other datasets.}
The results are shown in Fig.~\ref{fig.conv}.
We find \textit{NoisyTune} can consistently improve PLMs under different amounts of data, especially when less training data is used.
This is because the perturbed PLMs may have lower risks of overfitting the pretraining tasks and have better generalization abilities, which is especially beneficial for finetuning PLMs on downstream tasks with limited data.


\begin{figure}[!t]
  \centering
    \includegraphics[width=0.46\textwidth]{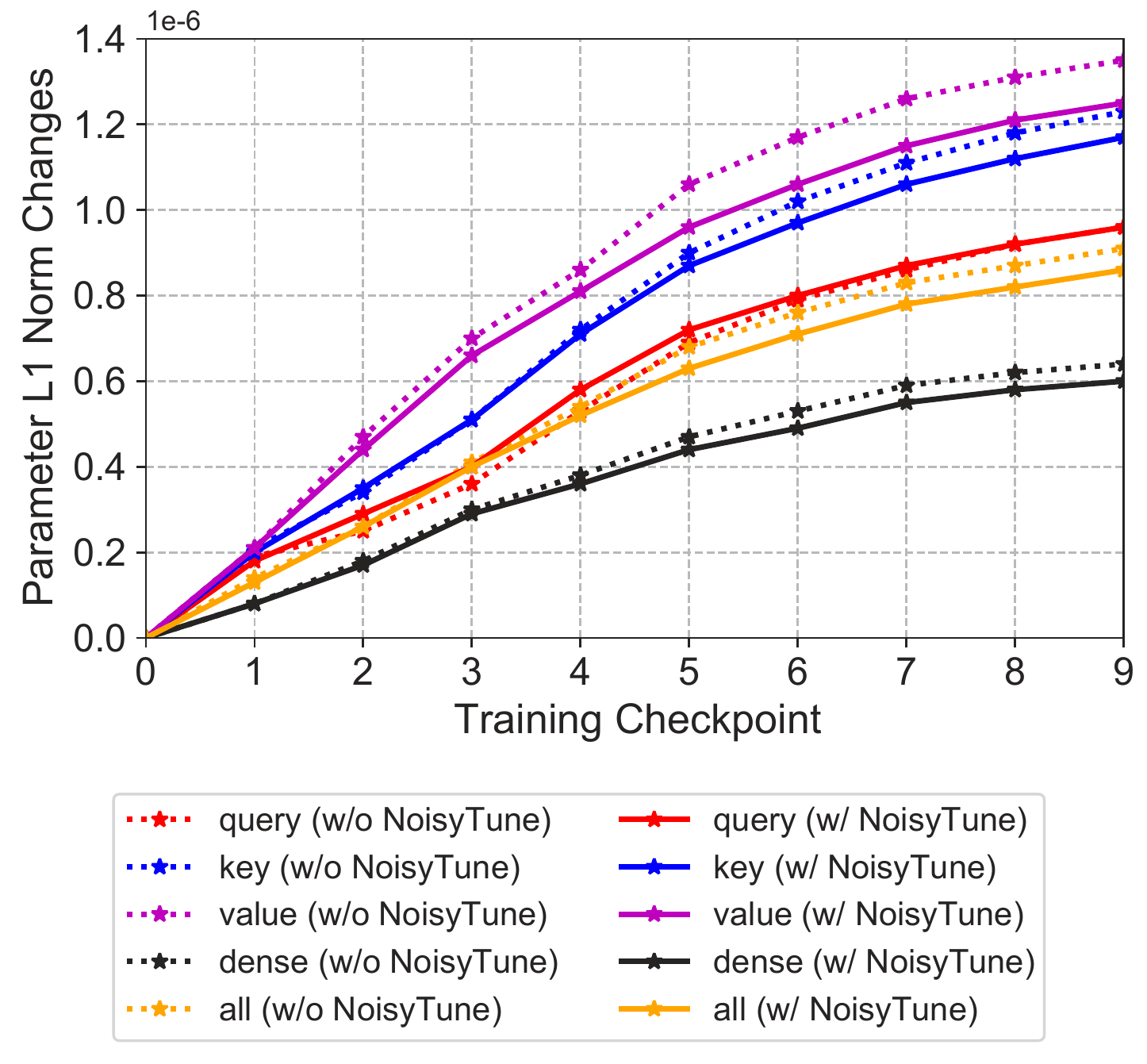}

  \caption{Relative changes of the L$_1$-norm of different types of parameters in PLM during finetuning.}\label{fig.norm}

\end{figure}

\begin{figure}[t]
  \centering
    \includegraphics[width=0.46\textwidth]{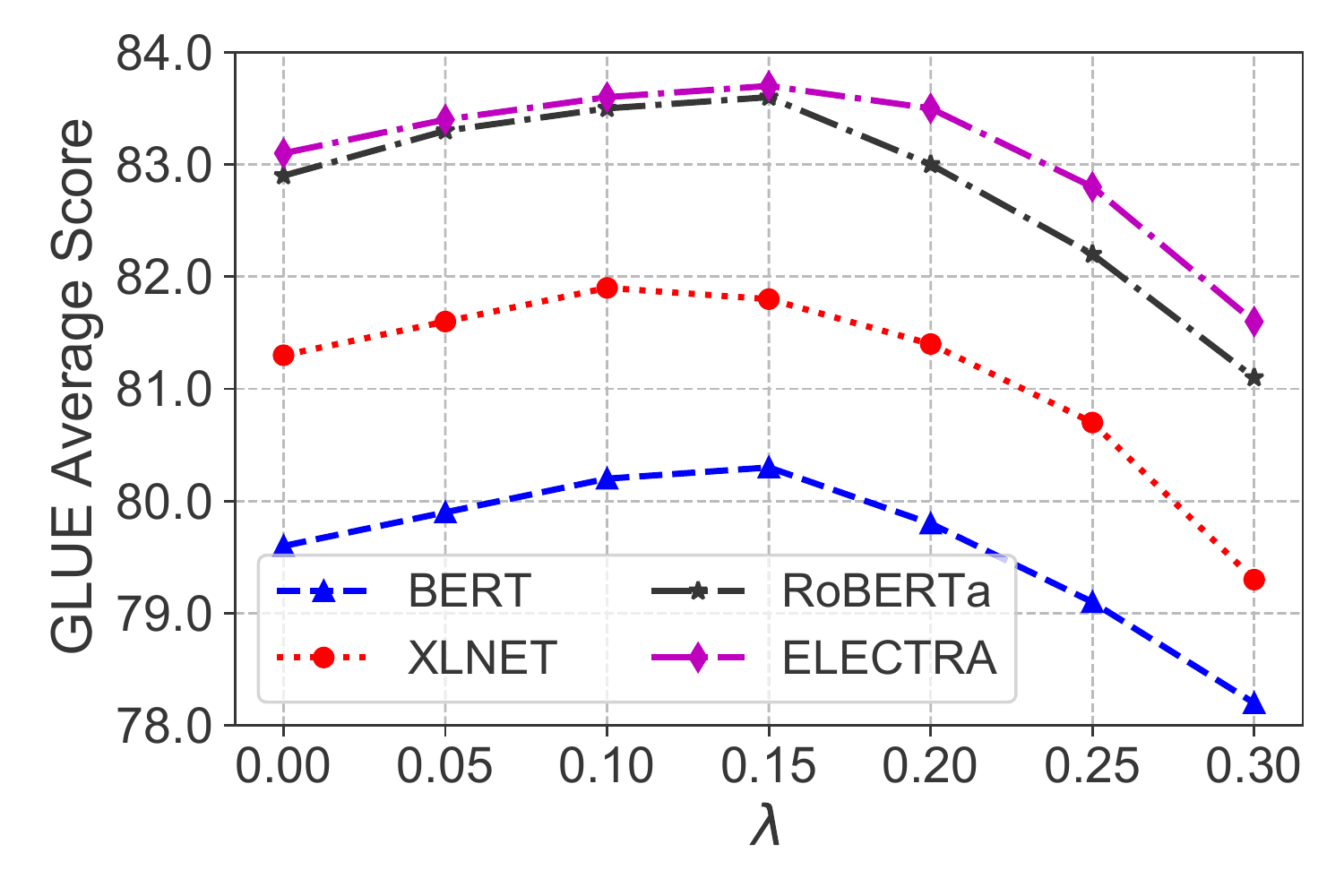}

  \caption{Influence of noise intensity $\lambda$.}\label{fig.scale}

\end{figure}

To further study the impact of \textit{NoisyTune} on PLM finetuning, we show the relative changes of the L$_1$-norms of different kinds of parameters in the BERT model during finetuning on the MRPC  dataset in Fig.~\ref{fig.norm}.\footnote{The patterns on other datasets are similar.}
Since the noise we added to PLMs in \textit{NoisyTune} is zero-mean uniform noise, the absolute parameter L$_1$-norm will not change too much.
However, we can see that the relative change of L$_1$-norms becomes smaller when \textit{NoisyTune} is applied, which indicates that the PLMs can find the (sub)optimal parameters for downstream tasks more easily.
This result validates directly finetuning PLMs may need more updates to adapt to downstream tasks, which is due to the overfitting of pretraining tasks, and \textit{NoisyTune} can provide a simple way to alleviate this problem and help finetune PLMs on downstream tasks more effectively.

\subsection{Hyperparameter Analysis}

We study the influence of the most important hyperparameter in \textit{NoisyTune}, i.e., $\lambda$, which controls the relative noise intensity.
The average GLUE scores w.r.t. different $\lambda$ values are shown in Fig.~\ref{fig.scale}.
We find that when $\lambda$ is too small or too large, the performance is not optimal.
This is because when $\lambda$ is too small, it is difficult for PLMs to do parameter space exploration and overcome the overfitting problem.
While when $\lambda$ is too large, the useful pretrained knowledge in PLMs may be overwhelmed by random noise.
Values between 0.1 and 0.15 are more suitable for \textit{NoisyTune} on the GLUE datasets.

\section{Conclusion}\label{sec:Conclusion}

In this paper, we propose a very simple but effective method named \textit{NoisyTune}, which can help better finetune PLMs on downstream tasks by adding a little noise to them before finetuning.
In \textit{NoisyTune}, we propose a matrix-wise perturbing method that adds noise with different intensities to different kinds of parameter matrices in PLMs according to their variances.
\textit{NoisyTune} is a very general method, and is PLM model agnostic, downstream task agnostic, and finetuning method agnostic.
Extensive experiments on both monolingual GLUE benchmark and  multilingual XTREME benchmark demonstrate \textit{NoisyTune} can consistently empower the finetuning of different PLMs on various downstream tasks to achieve better performance.

\section*{Acknowledgments}
This work was supported by the National Natural Science Foundation of China under Grant numbers U1936216, U1936208, and 61862002, and the research initiation project of Zhejiang Lab (No. 2020LC0PI01).

\bibliographystyle{acl_natbib}
\bibliography{acl2022}

\end{document}